\documentclass{article}
\usepackage{iclr2026_conference,times}
\usepackage{amsmath,amsfonts,bm}

\def\eqref#1{equation~\ref{#1}}
\def\1{\bm{1}}

\DeclareMathAlphabet{\mathsfit}{\encodingdefault}{\sfdefault}{m}{sl}
\SetMathAlphabet{\mathsfit}{bold}{\encodingdefault}{\sfdefault}{bx}{n}

\usepackage{hyperref}
\usepackage{url}
\usepackage{booktabs}
\usepackage{amsmath,amssymb}
\usepackage{graphicx}
\usepackage{multirow}
\usepackage{longtable}

\title{Emergence, Not Bandwidth: Physical Coupling and
the Limits of Learned Multi-Agent Communication}

\author{Mihir Chauhan \& Aniket Bera \\
IDEAS Lab, Department of Computer Science \\
Purdue University \\
\texttt{\{chauhanm,aniketbera\}@purdue.edu}
}

\iclrfinalcopy  % uncomment only for the camera-ready; submissions must stay anonymous

\begin{document}
\maketitle

\begin{abstract}
Rate-limited multi-agent teams raise three questions the emergent-communication
literature has answered only empirically: what an optimal message should encode,
what compression costs over a horizon, and when a learned protocol is unique enough
to be read by a teammate. We answer them for rate-limited Dec-POMDPs, and then
measure how far reinforcement learning falls short of the optimum the theory
locates. Our theorems fix what is achievable independently of any learner, so
a gap between an engineered sender and a learned one at the same bit budget is an
optimization fact rather than an information-theoretic one. We instantiate this on
three MuJoCo continuous-control arenas spanning zero, partial and rigid physical
coupling, with every condition charged exactly $2$ bits per decision by construction,
and the discriminating regime is produced by closing a physical side channel within
one arena, holding bodies, task and reward fixed. Communication value is governed by coupling: where the agents are
rigidly coupled through a shared object, no channel beats silence
($+0.001 \pm 0.001$, $p = 0.982$, $n = 25$), because proprioception already carries
what a message would say; where they are uncoupled, every condition solves the task; and
in the partially coupled regime the engineered $2$-bit sender reaches an
interquartile mean of $1.000$ while the learned $2$-bit sender reaches $0.482$ and
is indistinguishable from silence ($p = 0.400$, $n = 25$). Because the two senders
hold the same alphabet, bandwidth cannot explain the gap. Initializing a learned run from an engineered run's
receiver localizes the failure: the same channel then reaches $0.857$ against $0.562$
cold-started ($p < 0.001$), so the failure is neither representational nor a matter of maintenance: what reinforcement learning cannot do here is discover the protocol in the first place.
Cross-play shows the learned protocols are individually meaningful and mutually
unintelligible: self-play $0.980$ collapses to $0.144$ across seeds, and the best
alignment we can construct leaves at least $77\%$ of that gap standing. Every headline result
is reported at $25$ seeds per arena, against seven published baselines
reimplemented in our trunk at matched rate.
\end{abstract}

\section{Introduction}

Any team of robots operating under partial observability, where one agent may not observe what its partner observes, must either communicate or coordinate implicitly through the physics they share. Which of the two happens is usually treated as a property of the task. We argue that it is instead a property of \emph{physical coupling}, and that coupling can be varied as an independent variable while everything else is held fixed.

We use \emph{emergence} throughout in one narrow, behavioral sense: the event that end-to-end reinforcement learning, starting from a random initialization, arrives at a protocol that the receiver can act on. We do not use it to imply higher-order statistical structure, and we do not infer it from an information-theoretic signature. It is measured by task success at a fixed bit budget against an engineered sender that holds the same alphabet, so the claim that emergence fails is falsifiable by a learner that closes that gap.

This is important because emergent-communication literature has, almost entirely, studied only one end of that axis. The continuous-control work we found operated in multi-agent MuJoCo, which decomposes a \emph{single} robot body into agents controlling disjoint joint subsets. That construction is maximally, rigidly coupled. Our own measurements say it is the exact regime in which communication is a tight null: on a rigidly coupled bi-humanoid arena, an oracle channel, an engineered symbol, and a learned communication code are all within $0.001$ of silence. A literature clustered there is measuring the one setting whose answer is known in advance.

The regime that differentiates is the partially coupled one, where two agents that share an object but cannot both observe the target are coupled enough that their actions interact and uncoupled enough that communication helps. This is where our central result lies.

\paragraph{The theory--experiment relationship.} We use theorems to make an empirical point, following the structure of
\citet{dorner2025limits}: establish a bound that holds independently of any particular method, then show that practice falls short of even that bound. Our theorems characterize the optimum of a rate-limited Dec-POMDP without reference to a learner. When an engineered sender attains a level at 2 bits that a learned sender does not attain with the same 2 bits, the difference cannot be attributed to channel bandwidth; it is a statement about the optimization problem reinforcement learning has to solve.

\paragraph{Contributions:}
\begin{enumerate}
\item \textbf{Coupling as an independent variable} (Section~\ref{sec:arenas}). Three
continuous-control arenas: 1) uncoupled, 2) partially coupled, 3) rigidly coupled, each with shared networks, trainer, channels and rate accounting. Communication value is ordered by coupling, and the rigid regime is a null.
\item \textbf{A rate-matched engineered-versus-learned comparison}
(Section~\ref{sec:rate}). Every experiment's rate --- how many bits one agent can send per decision --- is derived from its channel configuration, not from realized message statistics. The engineered sender and the learned $4$-code channel are both $2$ bits per decision, as are three of
the published baselines, so the comparison is capacity-controlled.
\item \textbf{The separation itself} (Section~\ref{sec:main}). In the partially
coupled arena, the engineered $2$-bit sender outperforms silence by $+0.441 \pm 0.129$
($p < 0.001$) while the learned $2$-bit sender does not outperform it at all
($+0.068 \pm 0.163$, $p = 0.400$), and the direct matched-rate contrast between the
two is $+0.374 \pm 0.125$ ($p < 0.001$, $d_z = 1.23$).
\item \textbf{The failure is discovery} (Section~\ref{sec:warmstart}). Warm-starting
a learned run, where we load the parameters from an engineered receiver, rescues it ($+0.295 \pm 0.131$, $p < 0.001$), which rules out representational and maintenance failures and leaves the sender's search as the bottleneck.
\item \textbf{Protocol idiosyncrasy} (Section~\ref{sec:crossplay}). Learned
protocols that each solve the uncoupled arena are mutually unintelligible, with a cross-play gap of $0.836$ $[0.807, 0.862]$ of which the best alignment we can
construct recovers at most a quarter, bounded over all $600$ ordered seed pairs.
\item \textbf{Statistical scale.} $25$ seeds per arena and seven published
baselines, all under fixed arena, rate and seed set, with seed-paired tests, Holm correction, one-sided bounds in place of bare nulls, and interquartile
means with seed-level bootstrap intervals as recommended by
\citet{agarwal2021deep}.
\end{enumerate}

\section{Related work}

\paragraph {Emergent communication.} Learned protocols frequently fail to arise at all, or degenerate when they do. \citet{kottur2017natural} show that agents
solve referential tasks with protocols that are neither compositional nor interpretable. \citet{lowe2019pitfalls} show that sender-consistency metrics can be
high while the channel has no causal effect on the receiver. \citet{eccles2019biases}
attribute the failure to a joint exploration problem and propose positive-signaling
and positive-listening biases to rescue it. \citet{chaabouni2019anti} show learned
codes are inefficient without explicit length pressure. Our contribution is not the observation that emergence fails; it is the controlled measurement of \emph{what} fails, at a fixed bit budget, as a function of physical coupling.

\paragraph{The nearest neighbor, and why it does not settle the question.}
\citet{hill2025communicating} compare an engineered sender against a learned one in
gridworld task allocation and find the engineered protocol far stronger. The comparison is not rate-matched, and the asymmetry in message space is severe: their learned sender uses ``a binary message space $\{0,1\}$'' (1 bit per step) while
their engineered ``intention'' message is a continuous fixed-length vector produced
by attention over an imagined trajectory, with an unknown dimension. The engineered side is therefore unquantified, and a gap measured across that difference is fully consistent with the learned channel simply being too narrow. Our comparison removes that explanation by construction: both senders hold exactly
$4$ symbols with a fixed $2$ bits (Table~\ref{tab:rate}).

\paragraph{Discrete channels and bandwidth accounting.} Vector quantization
\citep{vandenoord2017neural} is the standard discrete bottleneck, and a line of work controls communication bandwidth explicitly \citep{wang2020imac,kapoor2025ddcl}.
These compare learned variants to other learned variants, but do not introduce a hand-engineered sender at a matched bit count, which is the comparison that differentiates bandwidth capacity from optimization. \citet{vanneste2022discretisation} benchmark
discretization estimators (DRU, ST-DRU, straight-through) and find some fail entirely in some environments; we include their estimators as baselines.

\section{Method}

\subsection{Setting and theory}

We formulate the problem as a Dec-POMDP with two agents. One agent observes a task-relevant variable (while the other does not) and may transmit a message over a channel limited to $R$ bits per
decision. We summarize the three results that characterize the optimum. Appendix~\ref{app:theory} gives
the full statements, the assumptions each requires, and the exact tabular validation.

\paragraph{Why the theory is important.} Our experiments show an engineered sender beating a learned one at an identical bit budget, which by itself is consistent with two bits simply not being enough. The engineered sender rules that out by construction, being a witness that two bits suffice to reach its level, but it does not explain \emph{why} the learned sender stops short. The theory supplies that reason. We state three theorems, T1, T2 and T3.

T1 fixes, independently of any learner, what an optimal rate-$R$ message must be a
statistic of: the receiver's advantage, not the sender's observation. This is what
makes a $2$-bit sector code a principled reference, and it
yields a prediction we can test on the emergent code itself, by tracking the receiver's steering state and \emph{not} the fixed sector label
(Appendix~\ref{app:semantics}). T2 bounds how compression loss scales with horizon, so a shortfall cannot be attributed to accumulated per-step error under filter
stability. T3 predicts that independently trained optimal protocols are mutually
readable after per-context relabeling, which Section~\ref{sec:crossplay} tests
directly and finds false of the protocols PPO actually produces.

The theorems convert a gap between two particular senders into a
falsifiable statement about the search problem; two of the three are falsified
against our own learners. This structure follows \citet{dorner2025limits}: establish what binds any method, then show that practice does not reach it.

\paragraph{T1: Advantage Sufficiency:} The optimal message is a randomized function
of the receiver-facing sufficient statistic, which is a statistic of the \emph{receiver's
advantage}, not of the sender's observation. The zero-loss rate is the entropy of
that statistic and not of the state. Codes that reconstruct the state waste $47\%$ of communication value at a matched rate on our tests and need roughly twice the rate to recover it. A $2$-bit engineered sender serves as a meaningful upper reference with respect to these results.

\paragraph{T2: Horizon-free Compression Loss:} Under filter stability
(Dobrushin coefficient $\rho < 1$) the value lost to compression is $\Theta(\epsilon H)$ compared to $\Theta(\epsilon H^2)$ in worst case, with a matching lower bound, confirmed experimentally with measured exponents $1.08$
and $2.01$.

\paragraph{T3: Identifiability:} Under a binding rate, task diversity, and a
generic value-separation condition, the set of optimal protocols collapses to a
single orbit of per-context relabelings, which would make decoders consistent and
zero-shot interoperation possible. Exhaustive computation shows the genericity
condition is load-bearing: identifiability is non-monotone in both rate and task weight and fails in symmetric configurations. We report in
Section~\ref{sec:crossplay} that T3's empirical leg \emph{fails} on these arenas, and why the cause is the arenas rather than the theorem.

\subsection{Arenas: coupling as an independent variable}
\label{sec:arenas}

All three arenas are MuJoCo \citep{todorov2012mujoco} continuous-control tasks that are
compiled through MJX, trained with PPO \citep{schulman2017proximal} under matched
networks, trainer, channel implementations and rate accounting. In each, one agent
observes a hidden variable taking one of four values and the other must act on it.

Comparing across the three is a between-arena design, and the arenas differ in
embodiment, task and, most importantly, physical coupling, so a difference between them
is not by itself attributable to coupling. We therefore also vary coupling within a single arena. On the bimanual arena, for example, the two arms hold one bar, and the bar's pose
already reveals where the partner is steering: an agent that can see the object can
follow it and never needs a message. For the blind agent, we set \texttt{partner\_blind}, which removes the bar
pose, the bar velocity, the tilt and the partner's hand from one agent's observation
while leaving its own proprioception, the clock, the bodies, the physics, the reward
and the rate accounting unvaried. The flag closes a physical sensing channel without changing the task, which makes coupling a manipulated variable on the same hardware. Arena specifications are in Appendix~\ref{app:arenas}.

\begin{description}
\item[Uncoupled (UAV--UGV).] An aerial observer can view which of four sectors holds the
goal; a ground vehicle must drive there. The agents cannot physically interact, so coupling is
zero and the message is the only information path.
\item[Partially coupled (bimanual, blind).] Two arms manipulate a shared object and the
target is hidden from one of them. The arms interact through the object by friction alone, and contact forces do not disclose the target.
\item[Rigidly coupled (bi-humanoid, blind).] Two humanoids carry an object through fixed handles, which effectively constrain each body to the
other continuously, so proprioception is itself an information channel.
\end{description}

\subsection{Rate accounting}
\label{sec:rate}

Rates are computed from each channel's configuration and not from realized message statistics. An entropy-coded charge would award a collapsed codebook a lower apparent rate, rewarding the exact failure under study, and would not be available to the engineered sender.

\begin{table}[t]
\caption{Communication rate on every arena, derived from each channel's configuration
and never from realized message statistics. Every
finite charge is a fixed-length cost over the alphabet the channel actually uses. The
engineered sender and the learned \texttt{vq\_k4} channel are exactly matched, as are
three published baselines.}
\label{tab:rate}
\begin{center}
\begin{tabular}{lc}
\toprule
\textbf{condition} & \textbf{bits/decision} \\
\midrule
engineered (scripted sector symbol) & 2.000 \\
\texttt{vq\_k4} (learned, ours)     & 2.000 \\
\texttt{cat\_k4} (Gumbel straight-through) & 2.000 \\
\texttt{dru\_b2} / \texttt{stdru\_b2} & 2.000 \\
\midrule
\texttt{vq\_k2} / \texttt{vq\_k8} / \texttt{vq\_k16} & 1.000 / 3.000 / 4.000 \\
silence & 0.000 \\
oracle (unquantized) & excluded from the rate axis \\
\bottomrule
\end{tabular}
\end{center}
\end{table}

\begin{figure}[t]
\begin{center}
\includegraphics[width=\textwidth]{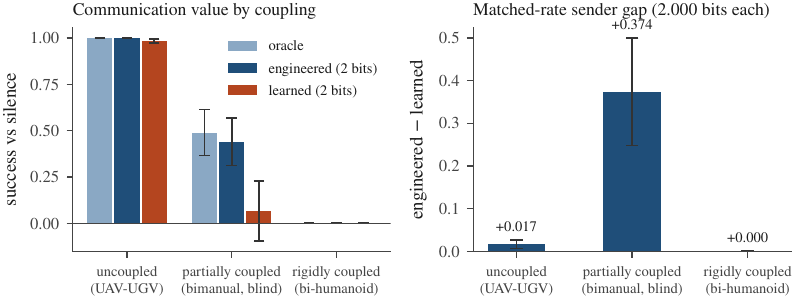}
\end{center}
\caption{Communication value is set by physical coupling. \textbf{Left:} each condition's
success against silence, $n = 25$ seeds per condition, $95\%$ $t$-intervals. \textbf{Right:}
the contrast of engineered minus learned, with both senders charged
exactly $2$ bits per decision. The matched-rate gap is negligible where the agents
are rigidly coupled (proprioception already carries the message), small where they are
uncoupled (any channel suffices), and large in between. Since the alphabets are
identical by construction, the middle bar is not a bandwidth effect.}
\label{fig:coupling}
\end{figure}

\subsection{Controls and nulls}
\label{sec:controls}

A gap between two senders can be caused by a weak baseline, a broken channel or a
favorable statistic, but we rule each out by our construction.

\begin{itemize}
\item \textbf{The floor is a strategy, not a failed learner.} The score under silence
$0.000$ on the uncoupled arena is a shaping equilibrium, so gaps there are reported
against the strongest scripted no-channel controller (Appendix~\ref{app:floor}).
\item \textbf{Rate is matched by configuration.} Both senders are allowed only $2$
bits per decision, so a gap cannot be bandwidth (Table~\ref{tab:rate}).
\item \textbf{The channel is not the weak link.} RIAL and ST-DRU beat our channel
where coupling is zero, but where it binds, neither closes the matched-rate gap.
\item \textbf{Physical coupling is manipulated.} The discriminating regime comes
from closing a physical sensing channel within one arena, holding bodies, task, reward and
rate fixed.
\item \textbf{Alignment is tested against a destroying null, which is bounded.} A shuffled control
recovers $0.086$ $[0.039, 0.137]$, so recovery above zero is not itself evidence of
alignment, and every non-significant contrast
carries a one-sided $95\%$ upper bound.
\item \textbf{Summaries resist outliers.} Means with $t$-intervals are reported
beside interquartile means with seed-level bootstraps
\citep{agarwal2021deep}.
\end{itemize}

\section{Results}

\subsection{Communication value by coupling regime}
\label{sec:main}

Figure~\ref{fig:coupling} shows the whole result; Table~\ref{tab:main} reports all three arenas. Seeds are paired across conditions, so each
contrast is a paired test on the same seeds; $p$-values are Holm-corrected within
each arena. We report the mean with a $95\%$ $t$-interval and, following
\citet{agarwal2021deep}, the interquartile mean with a seed-level bootstrap
interval. The two disagree on the partially coupled arena because its seed distribution
is bimodal rather than merely wide; Figure~\ref{fig:bimodal} shows every seed behind
that row.

\begin{table}[t]
\caption{Success against silence on three arenas, $n$ seeds per condition. IQM is the
interquartile mean with a seed-level bootstrap interval.}
\label{tab:main}
\begin{center}
\footnotesize
\setlength{\tabcolsep}{4pt}
\begin{tabular}{llccccc}
\toprule
\textbf{arena} & \textbf{condition} & $n$ & \textbf{success} & \textbf{IQM [95\% CI]} & \textbf{vs silence} & $p$ \\
\midrule
\multirow{4}{*}{uncoupled}
 & oracle      & 25 & $1.000 \pm 0.000$ & 1.000 [1.000, 1.000] & $+1.000 \pm 0.000$ & --- \\
 & engineered  & 25 & $0.999 \pm 0.001$ & 1.000 [1.000, 1.000] & $+0.999 \pm 0.001$ & $<0.001$ \\
 & \texttt{vq\_k4} & 25 & $0.983 \pm 0.010$ & 0.992 [0.977, 0.999] & $+0.983 \pm 0.010$ & $<0.001$ \\
 & silence     & 25 & $0.000 \pm 0.000$ & 0.000 [0.000, 0.000] & reference & \\
\midrule
\multirow{4}{*}{\textbf{partial}}
 & oracle      & 25 & $0.984 \pm 0.032$ & 1.000 [1.000, 1.000] & $+0.490 \pm 0.124$ & $<0.001$ \\
 & engineered  & 25 & $0.936 \pm 0.076$ & 1.000 [1.000, 1.000] & $+0.441 \pm 0.129$ & $<0.001$ \\
 & \texttt{vq\_k4} & 25 & $0.562 \pm 0.127$ & 0.482 [0.321, 0.719] & $+0.068 \pm 0.163$ & $\mathbf{0.400}$ \\
 & silence     & 25 & $0.494 \pm 0.124$ & 0.368 [0.297, 0.583] & reference & \\
\midrule
\multirow{4}{*}{rigid}
 & oracle      & 25 & $1.000 \pm 0.000$ & 1.000 [1.000, 1.000] & $+0.001 \pm 0.001$ & 0.982 \\
 & engineered  & 25 & $1.000 \pm 0.000$ & 1.000 [1.000, 1.000] & $+0.001 \pm 0.001$ & 0.982 \\
 & \texttt{vq\_k4} & 25 & $1.000 \pm 0.000$ & 1.000 [1.000, 1.000] & $+0.001 \pm 0.001$ & 0.982 \\
 & silence     & 25 & $0.999 \pm 0.001$ & 1.000 [1.000, 1.000] & reference & \\
\bottomrule
\end{tabular}
\end{center}
\end{table}

\paragraph{Coupling orders communication value.} In the rigidly coupled arena every
channel is within $0.001$ of silence and nothing is significant; the handles already
carry the information. In the uncoupled arena every channel solves the task, and
against the scripted floor of $0.242 \pm 0.037$ the learned sender gains
$+0.740 \pm 0.038$. Communication value is therefore not a property of the task
alone; it is set by how much the physics already couples the agents.

\paragraph{The matched-rate contrast.} We highlight the comparison between two
senders at the same rate. Paired on the $25$ seeds both conditions ran,
the engineered sender exceeds the learned one by $+0.374 \pm 0.125$ ($p < 0.001$,
Cohen's $d_z = 1.23$) in the partially coupled arena. The same contrast is
$+0.017 \pm 0.010$ ($p = 0.002$, $d_z = 0.68$) where the agents are uncoupled, and
exactly $+0.000 \pm 0.000$ where they are rigidly coupled. These three tests ask one
question of three arenas, so the two with defined $p$-values are Holm-corrected as a
family; the third is exactly zero with zero variance and admits no test. Correction
leaves both significant and the matched-rate gap is therefore not a fixed property of our
engineered protocol. It is zero under rigid coupling, small under no coupling, and
large in between.

\paragraph{What the null excludes.} The
one-sided $95\%$ upper bound on the learned channel's advantage over silence is
$+0.202$. The engineered sender's measured advantage on the same seeds is $+0.441$. The
learned sender's entire plausible range therefore lies below the engineered sender's
point estimate, and hence, a bound on the effect.

\paragraph{The separation.} The partially coupled arena is where the two senders come
apart at equal bandwidth. The engineered $2$-bit sender clears silence by
$+0.441 \pm 0.129$; the learned $2$-bit sender, holding the same alphabet under the same
charge, does not clear it ($p = 0.400$). The interquartile means sharpen this, as the 
engineered seed is at ceiling ($1.000$), while the typical learned
seed sits at $0.482$ against silence's $0.368$. The seed distribution here is bimodal,
which is why the IQM is the more faithful summary: the engineered mean of $0.936$ is
depressed by a few failing seeds, while the modal engineered seed solves the task outright.

\subsection{Published learners at the same rate}

Table~\ref{tab:baselines} places seven baseline communication learners on the
uncoupled arena under our trunk, arena, rate and seeds, differing only where their respective
papers differ. All clear silence decisively. Two estimators, RIAL and ST-DRU,
outperformed our channel by a small but significant margin, while IB-VQ, Eccles biases
and AE-grounding fall below it. Our channel is therefore not the strongest discrete
estimator available, and the claim does not depend on it being so. This arena instead
shows that a rate-matched discrete channel of any of these kinds succeeds where
coupling is zero.

\begin{table}[t]
\caption{Communication learners on the uncoupled arena. Same trunk, arena, rate and
seeds; paired on seeds, Holm-corrected across the table. This is a positive control, not
a ranking: all seven clear silence decisively and the whole field sits within $0.07$ at
the top of a saturated scale, which establishes that a rate-matched discrete channel of
any of these kinds succeeds where coupling is zero. The two that beat our channel here
are carried forward below to the arena that discriminates.}
\label{tab:baselines}
\begin{center}
\small
\begin{tabular}{llccc}
\toprule
\textbf{method} & \textbf{source} & $n$ & \textbf{success} & \textbf{vs ours} ($p$) \\
\midrule
DIAL + VQ (ours) & \citet{foerster2016learning} & 25 & $0.983 \pm 0.010$ & --- \\
RIAL             & \citet{foerster2016learning} & 25 & $0.998 \pm 0.003$ & $+0.016 \pm 0.011$ (0.024) \\
ST-DRU           & \citet{vanneste2022discretisation} & 25 & $0.998 \pm 0.002$ & $+0.016 \pm 0.009$ (0.010) \\
DRU              & \citet{foerster2016learning} & 25 & $0.991 \pm 0.005$ & $+0.009 \pm 0.011$ (0.105) \\
Eccles biases    & \citet{eccles2019biases} & 25 & $0.931 \pm 0.041$ & $-0.052 \pm 0.040$ (0.041) \\
AE-grounding     & \citet{lin2021grounding} & 25 & $0.931 \pm 0.043$ & $-0.052 \pm 0.045$ (0.050) \\
IB-VQ            & \citet{farooq2026ibvq} & 25 & $0.930 \pm 0.036$ & $-0.052 \pm 0.033$ (0.015) \\
\bottomrule
\end{tabular}
\end{center}
\end{table}

% BEGIN GENERATED estimator
% BEGIN GENERATED estimator --
\paragraph{The separation is not specific to our channel.} We suspected that
our VQ channel may simply be a weak estimator, so we test the two that outperformed ours
on the uncoupled arena, with the same $2$ bits. The two behave differently: RIAL reaches
$0.691 \pm 0.129$ ($n = 25$) and clears silence by $+0.196 \pm 0.150$ ($p = 0.013$), making it
the better estimator here, where ours does not clear silence at all. ST-DRU reaches
$0.486 \pm 0.117$ ($n = 25$) and does not clear it ($-0.008 \pm 0.138$, $p = 0.904$). However,
neither closes the matched-rate gap. The engineered sender exceeds RIAL by $+0.245 \pm 0.121$ and
ST-DRU by $+0.449 \pm 0.142$, both $p < 0.001$ after Holm correction. A stronger estimator
therefore recovers part of the channel's value and still leaves most of the gap standing,
which is what we expect if the obstacle is search rather than how the channel is written
down.
% END GENERATED estimator --

\subsection{Discovery, not representation or maintenance}
\label{sec:warmstart}

The separation in Section~\ref{sec:main} shows that a learned sender does not reach
what an engineered sender reaches at the same rate. However, this does not say which
part of the problem defeats it. Three failures are consistent with the evidence so far: the
channel may be unable to \emph{represent} a useful protocol, the optimizer may be
unable to \emph{discover} one, or it may discover one and fail to \emph{maintain} it.

We separate them by initializing a learned run from an engineered run's final
parameters. The engineered run spends its whole budget training a receiver to act on a
$2$-bit sector code and never trains a message head, because its symbol is scripted. A
warm-started run therefore begins with a receiver that already listens and a sender that
knows nothing. It asks the sender-side question directly: given a competent listener,
can PPO find a speaker? Parameter trees are identical by construction and the codebook
is always reinitialized, so nothing about the channel itself is inherited.

The transfer is not confined to the receiver, however: it carries both agents' motor
policies and the critic, and its $15$M steps come on top of the engineered run's $15$M. The
contrast below therefore bounds listener, motor initialization and budget together.

The answer is that it can on most seeds, and we see this on the paired $25$ seeds, where it improves on 16, is worse on 2 and ties on 7. The warm-started learned channel
reaches $0.857 \pm 0.096$ against the cold-started $0.562 \pm 0.127$, a difference of
$+0.295 \pm 0.131$ ($p < 0.001$, $d_z = 0.93$); Figure~\ref{fig:warmstart} joins the
two conditions seed by seed, and the gain is spread across a majority of seeds rather
than produced by a handful. It clears silence by
$+0.362 \pm 0.134$ ($p < 0.001$, $d_z = 1.11$), where the cold-started run does not
clear silence at all; it sits at $0.562$ against silence's
$0.494 \pm 0.124$. The engineered sender, at $0.936 \pm 0.076$, exceeds it by
$+0.079 \pm 0.092$ ($p = 0.090$, $d_z = 0.35$). This is not an equivalence, since a non-significant
difference is weak evidence either way, but the one-sided $95\%$ bound puts the
engineered advantage over the warm-started run at no more than $+0.155$, against
$+0.374$ over the cold-started one. Most of the gap closes when the sender is given
a listener.

From this it follows that: (1) the channel can represent a protocol that solves this task at two
bits, so the failure is not representational, and (2) the protocol also survives continued
training rather than decaying, so it is not a maintenance failure. Neither depends on the confounded inherited motor policies or the doubled step budget, and what end-to-end reinforcement learning cannot do here is find the protocol, which we show by elimination from (1) and (2), not from the $+0.295$
contrast, whose magnitude those two confounds inflate. This is the
joint-exploration account of \citet{eccles2019biases}, measured here at a fixed bit
budget against an engineered upper reference rather than inferred from a rescue
method.

This condition now runs the same $25$ seeds as every other, so
its cold, engineered, and silent references are the main table's rows rather than a
subset. The discovery contrast softened as seeds accumulated and then held --- $+0.372$
($d_z = 1.18$) at $n = 12$, $+0.272$ ($d_z = 0.83$) at $n = 16$, $+0.295 \pm 0.131$ ($d_z = 0.93$)
at $n = 25$ --- the ordinary behavior of an estimate early in a sweep, and the reason we
reported it throughout at the $n$ it had reached rather than the $n$ that flattered it.

\subsection{What the learned code carries}

On the uncoupled arena the learned protocol succeeds, so we can understand what messages mean.
Rolling out trained checkpoints under greedy actions and comparing the transmitted
code against candidate meanings, over $25$ seeds and $48$ episodes each, the code
carries the goal sector (lift $+0.176$ $[+0.141, +0.213]$, $p < 0.001$) and the
required steering bearing (lift $+0.167$ $[+0.137, +0.197]$, $p < 0.001$) in roughly
equal measure, with smaller but significant information about range ($+0.021$) and
elapsed time ($+0.011$), and nothing about whether the vehicle is already aligned
($+0.003$, n.s.). The emergent code is therefore neither a pure sector name nor a
pure control signal but a mixture of both. This is consistent with T1, under which the
optimal message is a statistic of the receiver's advantage rather than of the
sender's observation.

\subsection{Cross-play, and the limits of the identifiability test}
\label{sec:crossplay}

T3 predicts that optimal protocols form a single orbit of per-context relabelings,
which would make protocols from independent seeds mutually readable after alignment.
We tested this on the uncoupled arena across $25$ seeds, evaluating every one of the
$600$ ordered pairs of distinct seeds on held-out episodes, with seed-level
cluster-bootstrap intervals.

The protocols are individually excellent and jointly unreadable. Self-play success is
$0.980$ $[0.967, 0.991]$, while cross-play with no alignment is $0.144$
$[0.123, 0.171]$, a gap of $0.836$ $[0.807, 0.862]$. The question is whether any
relabeling recovers that gap, and none comes close. A global relabeling recovers only
$0.173$ $[0.126, 0.226]$ of it, and the per-context $\hat{G}$ alignment
recovers $0.147$ $[0.102, 0.196]$. These intervals exclude zero at this sample size, so
we state the result as a bound rather than a null: with $95\%$ confidence the best
alignment we can construct leaves at least $77\%$ of the cross-play gap standing.
Figure~\ref{fig:crossplay} draws every ordered pair, one panel per alignment map: a
bright diagonal against a dark field in all of them.

A bound is the right form for this claim because a \emph{shuffled control}, which
should destroy alignment structure entirely, itself recovers $0.086$ $[0.039, 0.137]$.
Recovery above zero is therefore not evidence of alignment. Some of it is an artifact of
relabeling into a better-occupied part of the codebook, and the shuffled interval
overlaps both real alignments from below. What separates the conditions is small, and it
does not point the way T3 predicts.

\paragraph{Why T3's empirical leg fails.} The direct test is
$\hat{G}$ minus global: $-0.022$ $[-0.043, -0.001]$, $P(\leq 0) = 0.983$. The
per-context alignment is significantly worse than the global one. At
$n = 8$ this same contrast was $+0.005$ $[-0.045, 0.047]$ and we reported it as
underdetermined; at $n = 25$ it resolves against the prediction. The reason for this
is occupancy; the four context cells hold $[1270.5, 484.5, 8.8, 1.6]$
time-paired steps on average because episodes terminate on success and almost
nothing survives into the later cells. $\hat{G}$ is granted one permutation per cell,
but two of those four permutations are estimated from $\sim9$ and $\sim2$
steps respectively, leaving $2.1$ distinct per-cell permutations per pair. These extra
degrees of freedom are hence fitted to noise, which costs variance and buys no signal ---
exactly the sign and the size of the deficit we measure. We tried a second
context partition and obtained the same degeneracy. The result is therefore
not evidence against T3 as a theorem, which we verify exactly at the tabular level; it is
evidence that these arenas cannot test it, and with enough power to say that the
per-context refinement actively hurts rather than merely failing to help.

\paragraph{Individually meaningful, collectively unreadable.} A decoder from code to
sector fitted within a single seed reads $0.535$ against a majority-class rate of
$0.445$. Transferred across seeds, the same decoder reads $0.271$ unaligned and $0.416$
after alignment, which is at or below chance. Each seed's code genuinely carries the
sector, and no other seed can read it. This is a stronger statement than protocol
sub-optimality: the learned solutions are idiosyncratic in a way that no relabeling we
can construct repairs.

\section{Discussion and limitations}

Three findings organize the picture: coupling sets what a channel is worth, the
matched-rate shortfall belongs to the sender's search rather than to the channel, and the
protocols that do emerge are mutually unreadable. We state the scope of each plainly.

\begin{itemize}
% BEGIN GENERATED learner_limitation --
\item \textbf{One optimizer family.} Every result uses PPO at $15$M environment steps
per seed. Within that family we do vary credit assignment: MAPPO's centralized critic
also fails to close the matched-rate gap (Appendix~\ref{app:learner}, $n = 11$). No
non-PPO learner, optimizer setting or longer budget was tested, and that remains the
most important open item.
% END GENERATED learner_limitation --
\item \textbf{T3's empirical leg fails on these arenas.} The per-context alignment
is significantly \emph{worse} than the global one, and the cause we identify ---
degenerate context occupancy, two of four cells nearly empty --- is consistent with
the measurement but not proved by it. Testing T3 properly needs an arena with
balanced occupancy, which we have not built.
\item \textbf{Our channel is not the strongest estimator.} RIAL and ST-DRU
significantly outperform our VQ channel on the uncoupled arena. We report this rather
than promote the strongest condition to ``ours''.
\item \textbf{The open-channel condition is unmeasured.} The side channel is closed by a
flag, so the arena can be run with it open; currently the manipulation is a design control, not a measured contrast.
\item \textbf{Three arenas from one family.} All three are MJX continuous-control
tasks. They dissociate along coupling, the axis of interest, but do not span
benchmark suites.
\end{itemize}

Readings below $15$M environment steps are excluded throughout: the codebook sits at
perplexity $1.00$ until roughly $2$M steps, and a channel that looks like flat failure at
$0.38$ reaches $0.98$ by $13$M.

\section{Conclusion}

Whether a team benefits from an explicit channel is decided by how strongly its members
are already coupled through physics, and the field has worked mostly where the answer is
``not at all''. Where it discriminates, two senders holding the same two bits diverge
sharply: the obstacle is not capacity but what reinforcement learning can find.

Matching the rate is what licenses that reading. The engineered sender is charged the
same two bits and still reaches the task, so the budget demonstrably suffices and the
shortfall cannot be a bandwidth effect. Warm starting narrows it further, to discovery
rather than representation or maintenance, while the theorems fix the reference point independently of any learner, which leaves the residue as a fact about optimization.
Work in this area should therefore report where it sits on the coupling axis: a result
obtained where the message is the only information path need not survive where the
bodies already talk.

\section*{LLM Disclosure}

We used a large language model for: research execution and writing assistance. An LLM-based coding assistant was used to aid in debugging and implementing parts of the experimental infrastructure, such as the job supervisor, parallelization of experimentation, and to aid in designing the controls, such as working out that a rate charged from realized message statistics would reward codebook collapse, and that the cross-play test must align per context rather than globally. The research question, arena design, theoretical results, and every claim are the authors' own. For writing assistance, an LLM was used to refine the manuscript: assist with mathematical proofs in Appendix \ref{app:proofs}, improving transitions, and catching grammatical errors. All text was reviewed and edited by the authors, who take full responsibility for its content. No numbers were produced by a language model.

\bibliography{refs}
\bibliographystyle{iclr2026_conference}

\appendix

\section{Theorem statements and exact validation}
\label{app:theory}

The assumption labels below are the registry the three theorems share.

\begin{center}
\small
\begin{tabular}{@{}llp{0.62\textwidth}@{}}
\toprule
& \textbf{assumption} & \textbf{content} \\
\midrule
A1  & finite spaces      & State, action, observation and message spaces are finite. \\
A2  & filter stability   & $\rho = \sup_a (1 - \mathfrak{d}(T_a))(2 - \mathfrak{d}(Z_a)) < 1$. Used by T2's upper bound; its lower bound constructs families with $\rho \to 1$. \\
A3  & binding rate       & $R < R_0 := \inf\{R' : \delta^\star(R') = 0\}$: the budget is strictly below what reaching the envelope needs, so the constraint is active at every optimum. \\
A4  & $\kappa$-diversity & The task family $\mathfrak{E}$ is $\kappa$-diverse at rate $R$ for some $\kappa > 0$. \\
A5  & bounded rewards    & $\|r\|_\infty \le r_{\max} < \infty$, with $H < \infty$ whenever $\gamma = 1$. \\
A6  & common design      & The protocol, environment and prior are common knowledge, and common randomness is available to stochastic encoders. \\
A7  & reachability       & Every history in the stated spaces has positive probability under some feasible pair, so the conditional laws are defined everywhere. \\
A8  & channel synchrony  & The channel is memoryless and per-step, and delivers within the step. \\
A9  & reward genericity  & Rewards avoid the measure-zero tie sets on which distinct semantic partitions achieve equal value. A9$'$ is the strengthened no-mixing variant, assumed rather than derived. \\
A10 & finite alphabet, finite horizon & $H < \infty$ and the channel is a finite alphabet over noiseless broadcast with $|\mathbb{M}^i| \le 2^R$ --- the setting of every experiment reported here. \\
\bottomrule
\end{tabular}
\end{center}

\paragraph{T1.B (advantage sufficiency)} Fix an optimal $(\pi^\star, \mu^\star)$ at
rate $R$. Define the receiver-facing statistic $\Psi^i_t := \sigma(\kappa^i_t,
\tilde{u}^i_t)$, where $\kappa^i_t$ is the receiver-posterior map and $\tilde{u}^i_t$ is
the response-profile map built from the continuation payoff net of downstream
information costs at the dual optimizer. Then for every dual optimizer
$\lambda^\star$ and every $(i,t)$ with $\lambda^\star_{i,t} > 0$,
$$I\big(H^i_t \,;\, M^i_t \mid \Psi^i_t,\, U\big) = 0.$$
The optimal message is hence a randomized function of a statistic of the receiver's
\emph{advantage}, not of the sender's observation, and the zero-loss rate is the
entropy of that statistic not the state. Corollary T1.C makes the state
reconstruction strictly suboptimal: in the tabular family
used for this check, the Hamming-optimal (state-reconstructing) code
achieves value $0.9064$ against an optimum of $1.0$, so $\delta = 0.0936$ of a total
communication value of $0.20$ --- $47\%$ wasted at a matched rate. This percentage,
we show, is a property of that family's parameters, not a universal constant.

\paragraph{T2 (horizon-free compression loss)} With $\rho = \sup_a (1 -
\mathfrak{d}(T_a))(2 - \mathfrak{d}(Z_a))$ the composite Dobrushin coefficient, and
under A1, A2, A5--A8 with broadcast topology: if $\rho < 1$, the total loss is
uniformly bounded in $H$ (including $H = \infty$) by
$\delta_{\mathrm{step}} / ((1-\gamma)(1-\gamma\rho))$, giving $\Theta(\epsilon H)$
rather than the $\Theta(\epsilon H^2)$ obtained by chaining per-step errors
naively. A matching lower-bound family shows that the quadratic rate is attained in the
worst case, so the dichotomy is tight rather than an artifact of the proof.

The exponent is fitted numerically on tabular families. In the \texttt{healing} family an
absorbing failure flag instead repairs with probability $\lambda$ per step, so
$\rho = 2(1 - \lambda)$ sweeps across $1$ as $\lambda$ varies, and the fitted exponent
moves with it (the \texttt{flag} rows are the same construction at $\lambda = 0$, where
the flag never repairs):

\begin{center}
\small
\begin{tabular}[t]{lcc}
\toprule
\textbf{family} & $\rho$ & \textbf{exponent} \\
\midrule
healing & 0.200 & 1.079 \\
healing & 0.500 & 1.097 \\
healing & 1.000 & 1.152 \\
healing & 1.600 & 1.356 \\
healing & 1.900 & 1.732 \\
healing & 1.980 & 1.946 \\
\bottomrule
\end{tabular}
\hspace{2.5em}
\begin{tabular}[t]{lcc}
\toprule
\textbf{family} & $\rho$ & \textbf{exponent} \\
\midrule
chain   & 0.088 & 1.071 \\
chain   & 0.875 & 1.082 \\
chain   & 1.137 & 1.086 \\
chain   & 1.662 & 1.096 \\
\midrule
flag    & 2.000 & 2.013 \\
\bottomrule
\end{tabular}
\end{center}

The stable regime sits at $\approx 1.08$ and the worst-case family reaches
$2.013$. Note that $\rho > 1$ is necessary but not sufficient for the quadratic
regime --- \texttt{chain}, a tracking task in which a target teleports to a uniform
cell with probability $\lambda$ and the sender's view of it is erased with probability
$\eta$, has $\rho$ above $1$ at some settings and stays near $1.08$ --- which is what a
one-directional bound predicts.

\paragraph{T3.1 (identifiability).} Assume A1, A5, A6, A7, A10 (model (a),
$H < \infty$), a task family $\mathfrak{E}$, A3 (binding rate $R < R_0$), A4
($\kappa$-diversity), and A9/A9$'$ (generic value separation and strict
tie-breaking). Then after trimming, $\overline{\mathrm{Opt}}(R) = \hat{G} \cdot
\mu^\star$: a \emph{single orbit} of the common-information-adapted relabeling group
$\hat{G}$. All elements share one semantic protocol, and the identifiability margin
$\eta$ is strictly positive.

The distinction between $\hat{G}$ and the constant relabeling group $G$ is
load-bearing and is why we align per context in Section~\ref{sec:crossplay}: with
$G$ the statement is false in general sequential problems. At $H = 1$ the two
coincide, which is why referential-game prior work does not see the difference.

Because T3.1 is a claim about the \emph{argmax set} of an optimization problem, it
can be computed exactly on small families rather than estimated --- no seed noise,
and no dependence on whether RL finds the optimum. Doing so shows A9 is not
removable: identifiability is non-monotone in the rate.

\begin{center}
\small
\begin{tabular}{lccccc}
\toprule
$K$ & \textbf{bits} & $R < R_0$? & $V^\star$ & \textbf{\#orbits} & \textbf{identifiable} \\
\midrule
2 & 1.000 & yes      & 0.9500 & 1 & yes \\
3 & 1.585 & yes      & 0.9750 & 2 & \textbf{no} \\
4 & 2.000 & no ($=R_0$) & 1.0000 & 1 & yes \\
\bottomrule
\end{tabular}
\end{center}

At $K = 3$ the rate still binds ($1.585 < R_0 = 2$) yet $\mathrm{Opt}(R)$ splits
into \emph{two} orbits. A binding rate constraint is therefore not by itself
sufficient for identifiability: the failure is an A9 violation, and identifiability
fails exactly at symmetric configurations. The checker also verifies
orbit-invariance (every relabeling of a protocol shares its value) for every
protocol enumerated, and reproduces mechanically each value
computed by hand for the parity family of Appendix~\ref{app:proof-t3}.

\section{Proof overviews}
\label{app:proofs}

The three results are proven in full in the supplementary material, with assistance from an LLM, as with the proofs in this section. 
This appendix gives
the argument of each at the level of its important steps and every overview names
the lemmas it compresses; the statements are checked exactly by enumeration on small tabular
families (Appendix~\ref{app:theory}).

\subsection{T1: optimal messages are sufficient statistics for receiver advantage}
\label{app:proof-t1}

Two preliminary repairs to the rate-limited model make the optimization well posed.
First, the rate functional must condition on the common randomness $U$: without that
conditioning the achievable region fails to be convex, and time sharing between two
feasible protocols can appear to violate the budget. Second, with the conditioned
functional, the region is convex and compact, so an optimum is attained and strong
duality holds. The consequence we use throughout is an exact Lagrangian bridge: there
is a multiplier vector $\lambda^\star$ with zero duality gap, so a maximizer of the
unconstrained $L_{\lambda^\star}$ that is also feasible is a constrained optimum. The
variational (Donsker--Varadhan) form of mutual information turns the rate term into a
sum of per-history stage costs $D(\mu(\cdot \mid h) \,\|\, q^\star)$ against a fixed
reference measure, which is what lets the information charge be treated as part of the
stage reward rather than as a global constraint.

\paragraph{T1.A (belief sufficiency).} Four steps. \emph{(1) Lagrangianization}: replace
the constrained problem by $L_{\lambda^\star}$, in which the rate appears as a
per-history cost. \emph{(2) Coordinator reduction}: under broadcast the common
information is shared, so the common-information approach of \citet{nayyar2013decentralized}
applies and an optimal solution exists in which each agent's stage kernel is a function
of the common belief and its own private information. \emph{(3) Belief quotient}:
two private histories inducing the same private belief are bisimilar for the
Lagrangianized problem --- they generate the same distribution over futures and the
same stage costs --- so the optimal kernel factors through the belief. \emph{(4)
Assembly}: the maximizer constructed in (1)--(3) is feasible, so by zero gap it is a
constrained optimum.

Broadcast is where this is fragile, and the fragility is specific. Step (2) needs every
agent to see the same public record; with point-to-point channels the common
information shrinks, the private belief no longer determines the common belief, and the
statement must carry the pair rather than the private belief alone. The quotient in
step (3) still goes through formally, so the failure is in the reduction, not the
bisimulation.

\paragraph{T1.B (advantage sufficiency).}  Fix an optimum
$(\pi^\star, \mu^\star)$ and a stage $(i,t)$ whose multiplier is strictly positive.
Define $\Psi^i_t := \sigma(\kappa^i_t, \tilde u^i_t)$ from the receiver-posterior map
and the response-profile map, the latter built from the continuation payoff net of
downstream information costs at the dual optimizer. Suppose, for contradiction, that
$I(H^i_t ; M^i_t \mid \Psi^i_t, U) > 0$.

Now garble: replace the stage kernel by
$\mu'(\cdot \mid h) := \mathbb{P}(M^i_t \in \cdot \mid \Psi^i_t = \Psi^i_t(h))$, drawing
the message from its conditional law given the atom of the statistic and nothing else.
Two facts make this an improvement. The receiver-facing joint law of
$(H^j_t, \tilde u^i_t, M^i_t)$ is \emph{exactly} preserved, because on an atom of
$\Psi^i_t$ the posterior map is constant, which gives
$H^j_t \perp H^i_t \mid \Psi^i_t$; every value term of the decomposition is therefore
unchanged, not merely bounded. And the information cost weakly drops, since the chain
rule gives $I(H^i_t ; M^i_t) = I(\Psi^i_t ; M^i_t) + I(H^i_t ; M^i_t \mid \Psi^i_t)$,
with the assumed strict positivity of the last term making the drop strict. So
$(\pi^\star, \mu')$ has the same value at strictly lower Lagrangian cost, contradicting
optimality. Hence the conditional information vanishes: the optimal message is a
randomized function of the receiver's advantage statistic, and the zero-loss rate is the
entropy of that statistic rather than of the state.

This is self-contained because the information costs are carried
\emph{inside} the stage payoff by the variational representation, so preserving the
receiver-facing law preserves them in expectation, which is all the Lagrangian
charges. This is why no assumption about the continuation being already optimal is
needed, and so why nothing is assumed that the conclusion supplies.

\paragraph{T1.C (state reconstruction is strictly suboptimal).} A code that is optimal
for reconstructing the sender's observation spends rate on distinctions the receiver's
advantage statistic does not separate. On the tabular family used for the exact check
this is $47\%$ of the available communication value at matched rate. The percentage is a
property of that family; T1.C asserts only the direction, and the direction follows from
T1.B because any state-reconstructing code with $I(H^i_t ; M^i_t \mid \Psi^i_t) > 0$ is
strictly improvable by the same garble.

\subsection{T2: a horizon-free bound when the filter contracts, and a quadratic one when it does not}
\label{app:proof-t2}

The object being bounded is the gap between the omniscient belief $\beta_t$, which
conditions on everything, and the deployed common belief $\hat\beta_t$ reconstructed
from what the rate-limited channel actually carried. Write $e_t := \mathbb{E}\|\beta_t -
\hat\beta_t\|_{\mathrm{TV}}$ and let $\varepsilon_t$ be the per-step message distortion
injected at $t$, with $\bar\varepsilon := \sup_t \varepsilon_t$.

\paragraph{The recursion.} The proof rests on a controlled one-step filter contraction.
Inserting the truth-updated deployed belief and applying the triangle inequality splits
the step-$t$ error into a propagated part and a freshly injected part:
$$\|\beta_t - \hat\beta_t\| \le \big\| F(\beta_{t-1}, A_{t-1}, O_t) - F(\hat\beta_{t-1},
A_{t-1}, O_t) \big\| + \big\| F(\hat\beta_{t-1}, A_{t-1}, O_t) - \hat\beta_t \big\|.$$
Conditioning on $\mathcal{G}_{t-1} := \sigma(O_{1:t-1}, U)$, the observation $O_t$ is
drawn from the $\beta_{t-1}$-side predictive, so the contraction lemma bounds the first
term in conditional expectation by $\rho\|\beta_{t-1} - \hat\beta_{t-1}\|$, where
$\rho = \sup_a (1 - \mathfrak{d}(T_a))(2 - \mathfrak{d}(Z_a))$ is the composite Dobrushin
coefficient of transition and observation kernels. The second term is $\varepsilon_t$ by
definition. Hence $e_t \le \rho\, e_{t-1} + \varepsilon_t$ with $e_1 \le \varepsilon_1$,
and unrolling gives $e_t \le \sum_{k \le t} \rho^{\,t-k} \varepsilon_k \le \bar\varepsilon
\min\!\big(t, 1/(1-\rho)\big)$. The dichotomy in the theorem is already visible here: the
minimum is $1/(1-\rho)$ when $\rho < 1$ and $t$ when it is not.

\paragraph{From belief error to value loss.} A performance-difference argument converts
$e_t$ into regret, with the per-step sensitivity $\kappa_t$ supplied by a Lipschitz and
greedy-mismatch bound. In the discounted case $\kappa_t \le r_{\max}/(1-\gamma)$ and
summing the geometric double sum gives
$J^\star - J \le \frac{2\kappa}{1-\gamma\rho}\sum_k \gamma^{k-1}\varepsilon_k$, which is
bounded uniformly in $H$ --- including $H = \infty$. In the undiscounted finite-horizon
case $\kappa_t = (H - t + 1) r_{\max}$, and the same sum yields
$r_{\max}\bar\varepsilon H(H+1)/(1-\rho)$: the extra factor of $H$ comes from the
sensitivity, not from the belief error, which is why a bounded-advantage condition
removes it and restores a linear rate.

\paragraph{Tightness.} The quadratic rate is attained, not merely permitted by the
bound. A flag family in which a rate deficit prevents the deployed belief from ever
re-synchronizing gives a matching $\Omega(\min(\varepsilon H^2, H))$ lower bound, so the
dichotomy is a property of the problem rather than slack in the analysis. The fitted
exponents in Appendix~\ref{app:theory} are the numerical counterpart: the stable regime
sits near $1.08$ and the flag family reaches $2.013$. One caveat the numbers make
visible is that $\rho > 1$ is necessary but not sufficient for the quadratic regime ---
the chain family exceeds $1$ at some settings and stays near $1.08$ --- which is exactly
what a one-directional bound predicts, and is a place where a reader could otherwise
suspect the fit rather than the theory.

Two honest limitations. The decoder must not rule out observations the true system can
produce; this is a support condition, and $\eta$-smoothing the deployed belief enforces
it at a cost of $2\eta$ in the distortion, so it is a technical rather than a
substantive hypothesis. And the claim that the deployed pair is admissible-and-feasible
is automatic in the models we use but is a hypothesis in general.

\subsection{T3: optimal protocols form a single orbit, under the right group}
\label{app:proof-t3}

T3.1 is a statement about the argmax set of an optimization problem, which is why it can
be verified exactly on small families rather than estimated: there is no seed noise and
no dependence on whether a learner finds the optimum.

\paragraph{The argument.} Five lemmas compose. Blockwise convexity and purification give
a deterministic optimum, and strict tie-breaking (A9$'$) upgrades this to: \emph{every}
optimum is deterministic on reachable histories. Trimming then leaves finitely many
deterministic protocols, hence finitely many \emph{semantic protocols} $\sigma(\mu)$ ---
the decision-relevant partition each protocol induces. Value factors through $\sigma$,
so the problem reduces to a finite maximization over semantic protocols; generic value
separation (A9) makes the maximizer $\sigma^\star$ unique. Finally, two deterministic
trimmed protocols share a semantic protocol exactly when they lie in one orbit of the
common-information-adapted relabeling group $\hat G$. Combining: every optimum has
semantics $\sigma^\star$ and therefore lies in $\hat G \cdot \mu^\star$, and the whole
orbit is optimal by $\hat G$-invariance. The margin $\eta > 0$ follows from finitely
many semantic values with a unique maximum; against cell-merging deviations,
$\kappa$-diversity (A4) gives the stronger bound $\eta \ge \kappa$ directly.

\paragraph{Why the group must be $\hat G$.} This is the load-bearing distinction and the
reason the paper aligns per context in Section~\ref{sec:crossplay}. With the constant
relabeling group $G$ the theorem is false in general sequential problems: the optimum
set is then a disjoint union of many $G$-orbits, because a protocol may permute its
labels differently at different common-information states without changing anything a
receiver can act on. At $H = 1$ the two groups coincide, which is why work on
referential games does not encounter the difference --- and why a cross-play experiment
that aligns with a single global permutation is testing a weaker statement than the one
T3 makes.

\paragraph{Necessity, and what the exact sweep adds.} A3 (a binding rate) and A4 are
close to necessary rather than convenient: if the rate does not bind, distinct protocols
achieve the optimum and identifiability fails; if $\kappa$-diversity fails, the margin
collapses. A9 is the assumption a reader is most likely to suspect of being
unfalsifiable, so the enumeration in Appendix~\ref{app:theory} is aimed at it: at
$K = 3$ the rate still binds ($1.585 < R_0 = 2$) and yet the optimum set splits into two
orbits. A binding rate is therefore \emph{not} sufficient for identifiability, the
failure occurs precisely at symmetric configurations where A9 is violated, and
identifiability is non-monotone in the rate. The same enumeration checks
orbit-invariance for every protocol it encounters and reproduces the hand-computed
values of the parity family, which is the construction underlying the interpretability
tax lower bound.

\section{What the emergent code carries}
\label{app:semantics}

Trained checkpoints on the uncoupled arena are rolled out under greedy actions and the
transmitted code is compared against candidate meanings, over $25$ seeds and $48$
episodes each. Lift is the accuracy of the best code-to-candidate lookup minus the
majority-class rate; intervals are bootstrap over seeds and $p$-values are
Holm-corrected across candidates. NMI is the mutual information between the code and
the candidate, normalized by the smaller of their entropies; because that denominator
depends on the candidate's own cardinality, NMI is not comparable across rows, and the
claims rest on lift.

\begin{table}[h]
\caption{The code carries the goal sector and the required steering bearing in roughly
equal measure. \texttt{aligned} --- whether the vehicle is already pointed correctly ---
is the one candidate it does not track.}
\begin{center}
\begin{tabular}{lcccc}
\toprule
\textbf{candidate meaning} & \textbf{NMI} & \textbf{lift} & \textbf{95\% CI} & $p$ (Holm) \\
\midrule
sector      & 0.232 & $+0.176$ & $[+0.141, +0.213]$ & $< 0.001$ \\
bearing (4 buckets) & 0.298 & $+0.167$ & $[+0.137, +0.197]$ & $< 0.001$ \\
turn sign   & 0.070 & $+0.058$ & $[+0.040, +0.077]$ & $< 0.001$ \\
range (4 buckets) & 0.029 & $+0.021$ & $[+0.011, +0.033]$ & 0.002 \\
elapsed time (4 buckets) & 0.035 & $+0.011$ & $[+0.006, +0.017]$ & 0.002 \\
aligned     & 0.218 & $+0.003$ & $[+0.000, +0.010]$ & 0.322 \\
\bottomrule
\end{tabular}
\end{center}
\end{table}

\section{The scripted floor on the uncoupled arena}
\label{app:floor}

The silent condition scores $0.000$, which is a shaping equilibrium rather than an
information bound: the policy learns to stand still and returns exactly
$-0.640 = 160 \times 0.004$, the horizon times the step cost. The honest floor is what
a strategy achieves with no channel, so we script one. It drives to a sector
center and, because the in-sector jitter ($0.45\sqrt{2} = 0.636$\,m) is smaller than the
sensing radius ($0.7$\,m), arriving reveals the goal whenever the guess was right.

Over $512$ episodes it reaches $0.242 \pm 0.037$. Time caps it: the horizon buys
$160 \times 5 \times 0.005 = 4.0$\,s at $1.6$\,m/s, so $6.4$\,m of travel, while the ring
hop between adjacent sectors is $1.9\sqrt{2} = 2.687$\,m. A second sector is reached in
$0\%$ of episodes, which is a real property of the arena and the reason it
discriminates at all. All gaps on this arena are reported against $0.242$ instead of silence.

\begin{figure}[h]
\centering
\includegraphics[width=\textwidth]{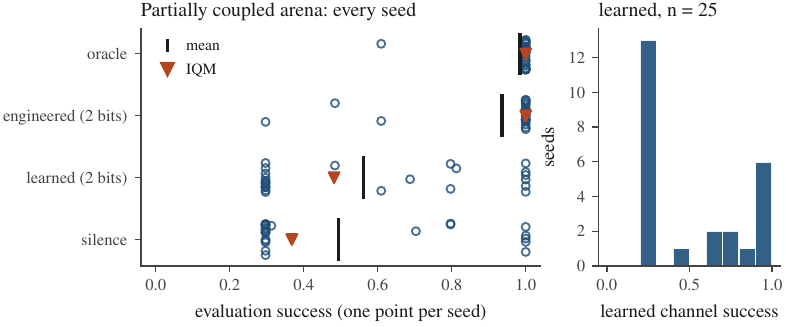}
\caption{Per-seed evaluation success on the partially coupled arena, one point per seed,
with the mean (bar) and interquartile mean (triangle) marked. The right panel histograms
the learned channel alone. The paper reports an interquartile mean beside the mean on this
arena because the distribution is bimodal; this is the evidence for that, and it also
shows why the learned channel and silence are hard to separate: they have the same
two-cluster shape, and the learned channel merely moves a few more seeds into the upper
cluster. Values are those of Table~\ref{tab:main}.}
\label{fig:bimodal}
\end{figure}

\begin{figure}[ht]
\centering
\includegraphics[width=\textwidth]{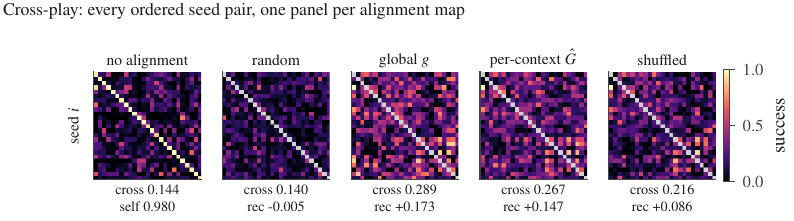}
\caption{Cross-play success for every ordered pair of seeds on the uncoupled arena, one
panel per alignment map. The bright diagonal in the leftmost panel is self-play; the dark
field is every other pairing. Diagonal cells are gray in the alignment panels because a
map from a seed to itself is not a pairing that was run. Panel labels give the cross-play mean and the share of the self-play gap the
map recovers, both read from the same data as Section~\ref{sec:crossplay}.}
\label{fig:crossplay}
\end{figure}

\begin{figure}[ht]
\centering
\includegraphics[width=0.62\textwidth]{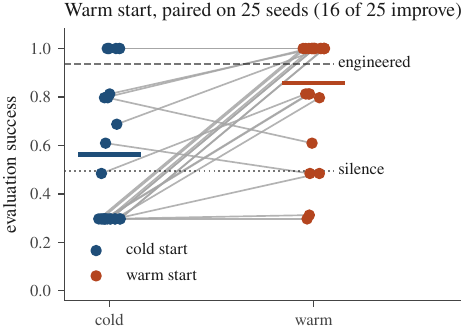}
\caption{Warm- against cold-started runs on the partially coupled arena, joined per seed,
with the engineered and silent references drawn as horizontal lines. The steep bundle is
the discovery effect: seeds that sit near the silent floor when cold-started reach the
engineered level when handed a competent receiver. Seeds that do not improve are visible
rather than absorbed into a mean. Values are those of Section~\ref{sec:warmstart}.}
\label{fig:warmstart}
\end{figure}

% BEGIN GENERATED learner --
\section{Does the failure survive a different learner?}
\label{app:learner}

Every headline number comes from IPPO at one configuration, so ``learned communication
fails here'' could be read as ``this optimizer fails here''. We re-run the partially
coupled arena under MAPPO, a centralized critic, which is a genuinely different
credit-assignment scheme rather than a retuning of the same one, with its own
engineered and silent references, so the contrast is internally paired.

\begin{center}
\begin{tabular}{@{}lccc@{}}
\toprule
critic & $n$ & learned & engineered $-$ learned \\
\midrule
IPPO (main text) & 25 & $0.562 \pm 0.127$ & $+0.374 \pm 0.125$ \\
MAPPO (centralized critic) & 11 & $0.531 \pm 0.191$ & $+0.447 \pm 0.206$ \\
\bottomrule
\end{tabular}
\end{center}

The centralized critic does not close the matched-rate gap ($+0.447 \pm 0.206$, $p = < 0.001$), and its
learned condition lands inside the IPPO interval. Sharing a critic changes how credit is
assigned, not whether the sender's search finds a protocol, which is what the warm-start
result already implied. This comparison is reported at the $n$ it reached.
% END GENERATED learner --

\section{Arena specifications}
\label{app:arenas}

Every quantity below is read from the environment configuration that the runs use.
All three arenas integrate at $5$\,ms with five physics substeps per control step, giving a
$25$\,ms control period ($40$\,Hz); actions are continuous and clipped to $[-1, 1]$. The
hidden variable is one of four values in each arena, drawn uniformly at reset, and is
visible to agent $0$ (the sender) only. Observation widths include the $8$-slot message
input. The \texttt{partner\_blind} flag zeroes the listed coordinates of the receiver's
observation rather than removing them, so widths and network shapes do not change when the
flag is set.

\begin{center}

\small
\setlength{\tabcolsep}{4pt}
\begin{tabular}{@{}lccc@{}}
\toprule
& \textbf{uncoupled} & \textbf{partially coupled} & \textbf{rigidly coupled} \\
& (UAV--UGV) & (bimanual, blind) & (bi-humanoid, blind) \\
\midrule
horizon (control steps)        & 160          & 120          & 150 \\
observation dim, sender        & 23           & 29           & 30 \\
observation dim, receiver      & 19           & 25           & 26 \\
\quad zeroed when blind        & ---          & 10 of 17     & 9 of 18 \\
action dim, sender / receiver  & 3 / 2        & 3 / 3        & 4 / 4 \\
\midrule
hidden variable                & goal sector  & target       & target zone \\
\quad cardinality              & 4            & 4            & 4 \\
target ring radius (m)         & 1.9          & 0.22         & 0.95 \\
success tolerance (m)          & 0.35         & 0.07         & 0.30 \\
lift requirement (m)           & ---          & 0.16         & 0.14 above start \\
\midrule
success reward (once)          & 1.0          & 1.0          & 1.0 \\
progress shaping (per m)       & 0.35         & 1.2          & 0.6 \\
lift shaping (per m)           & ---          & 0.6          & 0.9 \\
step cost                      & 0.004        & 0.004        & 0.004 \\
drop penalty                   & ---          & 0.25         & 0.30 \\
\bottomrule
\end{tabular}

\end{center}

{\textbf{Success criteria.} \emph{Uncoupled:} the UGV body comes within $0.35$\,m of the
goal, which sits at a jittered point ($\pm 0.45$\,m) on one of four sector centres at radius
$1.9$\,m; the UGV cannot sense the goal at all beyond $0.7$\,m, which is what makes the
message the only information path. \emph{Partially coupled:} the bar centre clears
$0.16$\,m and comes within $0.07$\,m in the $xy$ plane of the target. The hands begin
already gripping the bar (a $0.02$\,m inward preload, settled for $40$ physics steps), so
the task is coordinated transport rather than grasp discovery, though the grip can still be
lost. \emph{Rigidly coupled:} the box is carried $0.14$\,m above where it sits when control
starts and comes within $0.30$\,m of the target zone; a box roll or pitch beyond $0.55$\,rad
counts as a one-sided lift and ends the episode as a drop. Success is latched on first
attainment and rewarded once. Episodes end on success, on a drop, or at the horizon.}

\end{document}